\documentclass[runningheads]{llncs}
\usepackage{graphicx}
\usepackage{float}
\begin{document}
\title{Deep Transformer Model with Pre-Layer Normalization for COVID-19 Growth Prediction}
%
%
\author{Rizki Ramadhan Fitra\inst{1} \and
Novanto Yudistira\inst{2} \and
Wayan Firdaus Mahmudy\inst{3}}
\authorrunning{R. Author et al.}
%
\institute{Informatics Engineering, Faculty of Computer Science, Brawijaya University, Indonesia\\
\email{rizkiramadhanf@gmail.com} \and
Informatics Engineering, Faculty of Computer Science, Brawijaya University, Indonesia\\
\email{yudistira@ub.ac.id} \and
Informatics Engineering, Faculty of Computer Science, Brawijaya University, Indonesia\\
\email{wayanfm@ub.ac.id}}
\maketitle              
\begin{abstract}
Coronavirus disease or COVID-19 is an infectious disease caused by the SARS-CoV-2 virus. The first confirmed case caused by this virus was found at the end of December 2019 in Wuhan City, China. This case then spread throughout the world, including Indonesia. Therefore, the COVID-19 case was designated as a global pandemic by WHO. The growth of COVID-19 cases, especially in Indonesia, can be predicted using several approaches, such as the Deep Neural Network (DNN). One of the DNN models that can be used is Deep Transformer which can predict time series. The model is trained with several test scenarios to get the best model. The evaluation is finding the best hyperparameters. Then, further evaluation was carried out using the best hyperparameters setting of the number of prediction days, the optimizer, the number of features, and comparison with the former models of the Long Short-Term Memory (LSTM) and Recurrent Neural Network (RNN). All evaluations used metric of the Mean Absolute Percentage Error (MAPE). Based on the results of the evaluations, Deep Transformer produces the best results when using the Pre-Layer Normalization and predicting one day ahead with a MAPE value of 18.83. Furthermore, the model trained with the Adamax optimizer obtains the best performance among other tested optimizers. The performance of the Deep Transformer also exceeds other test models, which are LSTM and RNN.

\end{abstract}
\section{Introduction}
Coronavirus disease, abbreviated as COVID-19, is an infectious disease caused by the SARS-CoV-2 virus~\cite{WorldHealthOrganization2021}. Based on the timeline of the World Health Organization (WHO) response to cases of COVID-19~\cite{WorldHealthOrganization2021a}, this case was first announced at the end of December 2019. The Health Commission of the City of Wuhan, China, reported a number of pneumonia cases caused by infection. The cause of the infection was later identified as a new type of coronavirus, the SARS-CoV-2 virus. Based on COVID-19 data on Worldometer~\cite{Worldometer2021} as of September 20, 229,292,520 cases have been confirmed worldwide and 4,190,763 cases in Indonesia. The first confirmed case in Indonesia was on March 2, 2020, with as many as 2 cases. Meanwhile, on September 19, 2021, the latest daily cases amounted to 2,234 cases.

According to Dama and Sinoquet~\cite{Dama2021}, time series prediction is critical in many domains. For example, in business and finance, it is used to plan policies by various organizations. It is used in industry and digital transactions to detect abnormal or fraudulent situations. In meteorology, it guides the correct decision-making for agriculture, air navigation, and sea. In medicine, it is used to predict the spread of disease, estimate mortality rates, and assess time-dependent risks. Furthermore, time series predictions in the business sector can forecast the inflation rate~\cite{Mahmudy2021,Alfiyatin2019}. It can also be used in the agricultural sector to predict the season and rainfall~\cite{Sihananto2017,Wahyuni2018}.

The growth of COVID-19 cases can be predicted using several approaches and methods, one of which is the Deep Neural Network (DNN) model. According to Putra~\cite{Putra2019}, DNN is a Neural Network with many layers. Generally, it has more than three layers: an input layer, a hidden layer that is more than or equal to two, and an output layer. One of the DNN models is the Transformer model proposed by Vaswani et al.~\cite{Vaswani2017}. This model relies entirely on attention mechanisms and is used for language translation.

Various methods have been developed to predict time series using the Transformer model. Li et al.~\cite{Li2019} conducted a study that proposed the LogSparse Transformer model using the Transformer model by Vaswani et al.~\cite{Vaswani2017}, which initially aimed to translate language into a model that can predict time series. It was found that the LogSparse Transformer exceeds the performance of several models tested of Autoregressive Integrated Moving Average (ARIMA), Error Trend Seasonal (ETS), Temporally Regularized Matrix Factorization (TRMF), RNN-based autoregressive model (DeepAR) and RNN-based state space model (DeepState). Furthermore, Wu et al.~\cite{Wu2020} also developed the Transformer model into a Deep Transformer model for time series prediction. This study found that using the Deep Transformer model, the Pearson correlation value of 0.928 and RMSE 0.588 beat the ARIMA, Long Short-term Memory (LSTM) model, and Seq2Seq+attention. Subsequent research by Lim et al.~\cite{Lim2021} proposed a Temporal Fusion Transformer (TFT) model that can predict multi-horizon univariate time series and support temporal dynamic interpretation. This model was tested by comparing it with several models: ARIMA, ETS, TRMF, DeepAR, Deep State-Space Model (DSSM), ConvTrans, Seq2Seq, and Multi-horizon Quantile Recurrent Forecaster (MQRNN), the result is the proposed model outperforms other models in all test scenarios.

Then, research by Zerveas et al.~\cite{Zerveas2021} proposed a Time Series Transformer (TST) model to predict multivariate time series. In this study, time series regression and classification were carried out using several datasets, which were then compared with other models. The results of this study showed that the TST model outperformed the average evaluation value for time series regression and time series classification. Furthermore, Zhou et al.~\cite{Zhou2020} proposed the Informer, a long-sequence time-series forecasting (LSTF) model. The Informer model gets test results that exceed the LogTrans, Reformer, LSTMa, Long- and Short-term Time-series network (LSTnet) models, DeepAR, ARIMA, and Prophet.

Xiong et al.~\cite{Xiong2020} experimented on the Transformer model~\cite{Vaswani2017} by comparing the position of the normalization layer and the warm-up stage of the learning rate model. From the experimental results, the original Transformer model that puts the normalization layer outside the residual block gets the expected value of the gradient from the parameter near the output layer, which is large at initialization and causes unstable training when using a large learning rate. Furthermore, it is found that the Transformer model with the normalization layer position in the residual block can be trained without a warm-up stage and converges more quickly.

There are several studies conducted to predict positive cases of COVID-19. One of them is a study by Yudistira~\cite{Yudistira2020} which uses multivariate Long Short-Term Memory (LSTM) to predict the growth of positive cases of COVID-19 in the world. In this study, several test scenarios were carried out and the results showed that the LSTM model could exceed the performance of ordinary Recurrent Neural Network (RNN) and Vector Auto Regression (VAR). Then, Yudistira et al.~\cite{Yudistira2021} analyzed which variables contributed to the time series prediction in COVID-19 cases. These variables include demographics, economy, health, government, education, environment, and COVID-19. The analysis was carried out by making a saliency map using gradient-based visual attribution for each variable and its time dimension. The results of the test obtained a saliency map to be able to understand what factors affect the increase or decrease of cases.

Based on the research results described previously, no studies predict the growth in the number of COVID-19 cases in Indonesia using the Deep Transformer model. Experimental research that compares the position of the normalization layer has only been carried out on the original Transformer model, not yet on the Transformer for time series prediction. Experiments for hyperparameter optimization have also not been carried out by all the previously described studies. Therefore, this study uses the Deep Transformer model~\cite{Wu2020} to predict the growth in the number of COVID-19 cases in Indonesia, compare the position of the normalization layer to the predicted results, and optimize several hyperparameters. The results of this study are expected to be helpful in various scientific fields in making decisions during this pandemic.

In this research, several test scenarios will be carried out. First, we test to get the best hyperparameters from the Deep Transformer model. The hyperparameters are the size of the model, the number of encoder and decoder blocks, the dimensions of the Feed-Forward Network, Pre-Layer, and Post-Layer, and the amount of time lag. Then, after obtaining the best hyperparameters, several advanced tests were carried out: testing the predictions for the next few days, testing the number of input features, and finally comparing them with the Recurrent Neural Networks (RNN) and Long Short-Term Memory (LSTM).

This paper will be divided into several sections. Section 1 explains the introduction and background of conducting this research. Section 2 describes previous research that is similar to this study. Finally, section 3 describes the methods used in the research, starting with the normalization of the data used, the Deep Transformer model, Mean Absolute Percentage Error (MAPE) to calculate errors and related to the dataset.

\section{Related Work}
Research that has been done by Vaswani et al.~\cite{Vaswani2017} proposes a Transformer model that is used to perform language translation. The transformer is a model architecture that relies entirely on attention mechanisms to draw global dependencies between input and output. The Transformers enable significantly more parallelization and greatly improve translation quality after training for just twelve hours on eight P100 GPUs. The attention mechanism used is Scaled Dot-Product Attention and Multi-head Attention. Product Attention calculates the attention function on a set of queries simultaneously. Then it is entered into the matrix. Multi-head Attention allows the model to pay attention to information from different sub-spaces of representation at different positions.

Li et al.~\cite{Li2019} conducted a study using the Transformer model for time series prediction. The research obtained satisfactory results, but there are two shortcomings: locality-agnostic and memory bottleneck. The first shortcoming can be overcome by the convolutional self-attention method, during the second shortcoming by the LogSparse method, so the proposed model name is LogSparse Transformer. This model is then compared with several popular models for time series prediction. These are Autoregressive Integrated Moving Average (ARIMA), Seasonal Error Trend (ETS), Temporally Regularized Matrix Factorization (TRMF), RNN-based autoregressive model (DeepAR), RNN-based state, and Space model (DeepState). It is found that the proposed model can exceed the performance of other models.

Furthermore, Wu et al.~\cite{Wu2020} also conducted research that developed a Transformer model so that it can be used to make time-series predictions. The model is called the Deep Transformer. This model was evaluated and obtained a value of 0.928 for the Pearson correlation and 0.588 for the Root Mean Square Error (RMSE). Furthermore, this study found that the Deep Transformer model is superior to the ARIMA, Long Short-term Memory (LSTM) model, and Seq2Seq+ Attention.

Furthermore, Lim et al.~\cite{Lim2021} developed a Transformer model to predict multi-horizon univariate time series and support the interpretation of temporal dynamics. This model is called the Temporal Fusion Transformer (TFT). The TFT model was tested by comparing it with several models: ARIMA, ETS, TRMF, DeepAR, Deep State-Space Model (DSSM), ConvTrans, Seq2Seq, and Multi-horizon Quantile Recurrent Forecaster (MQRNN), the result is the proposed model outperforms other models in all test scenarios.

Furthermore, Zerveas et al.~\cite{Zerveas2021} conducted a study that developed the Transformer model into a Time Series Transformer (TST) model to predict multivariate time series. In this study, time series regression and time series classification were carried out using several datasets, which were then compared with other models such as Support Vector Regression (SVR), Random Forest, XGBoost, Rocket, LSTM, and several others. The TST model is divided into two tests: TST with supervised training and TST with unsupervised pre-training. The results of these tests showed that the average value of the evaluation for time series regression and time series classification was superior to the TST model with unsupervised pre-training and continued with TST with supervised training.

Furthermore, Zhou et al.~\cite{Zhou2020} proposed a Transformer-based model that can predict a long sequence of time series or Long Sequence Time-series Forecasting (LSTF) called the Informer. This model applies the ProbSparse self-attention mechanism. The performance of the Informer was tested by comparing it with several models: LogTrans, Reformer, LSTMa, Long- and Short-term Time-series network (LSTnet), DeepAR, ARIMA, and Prophet. From the test results, it was found that the Informer model almost surpassed all the models tested.

Xiong et al.~\cite{Xiong2020} experimented on the Transformer model by comparing the position of the normalization layer and the warm-up stage of the learning rate model. The experiment was carried out by combining the location of the normalization layer before or after the Transformer layer and using a warm-up stage, which is used or not on the translator machine. From the experimental results, the original Transformer model, which puts the normalization layer outside the residual block, gets the expected gradient value of the parameter near the output layer, which is large at initialization. Therefore, it causes unstable training when using a large learning rate. Furthermore, it is also found that the Transformer model with the normalization layer position in the residual block can be trained without a warm-up stage and converges more quickly.

Based on previous studies, it can be said that the Transformer model produces a time series prediction model with robust performance. This study conducts time-series predictions using the Deep Transformer model on Indonesia's daily growth dataset of COVID-19 data. Moreover, we assume that the position of the normalization layer influences the optimal parameters generated during learning.

\section{Methods}
\subsection{Min-Max Normalization}
Min-max normalization according to Patro and Sahu~\cite{Patro2015} is a scaling technique or mapping technique with the aim of creating a new range specifically with predetermined limits. Min-max normalization can be calculated using the formula in Equation \ref{eq:minmax}. Variable A' is the min-max normalized data, A is the original data, D and C are the lower and upper ranges of the specified limits or can be written as [C,D].
\begin{equation}
\label{eq:minmax}
A^\prime=\frac{A-\min value\ of\ A}{\max value\ of\ A-\min value\ of\  A}*(D-C)+C 
\end{equation}

\subsection{Deep Transformer Model}
\begin{figure}
\centering
\includegraphics[width=8cm]{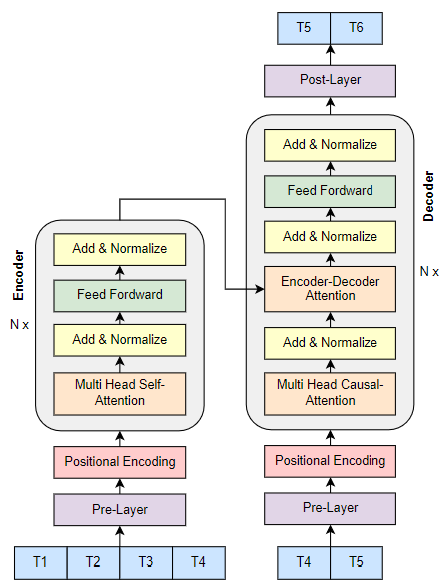}
\caption{Deep Transformer Architecture.} \label{architecture}
\end{figure}

Based on previous research by Wu et al.~\cite{Wu2020}, the architecture of the Deep Transformer model follows the original architecture of the Transformer by Vaswani et al.~\cite{Vaswani2017}. Deep Transformer consists of two layers, those are encoder and decoder. Unlike the original architecture, in this model the input and output embedding layers are changed to input layers, which are fully connected layers (see
Fig.~\ref{architecture}).

\subsubsection{Encoder}
The encoder consists of an input layer, a positional encoding layer, and a stack of four identical encoder layers. At the input layer, input in the form of time series data is mapped into ${d}$ dimensional vector model through a fully-connected network. In this step, a multi-head attention mechanism is carried out which is an essential step for the model. The positional encoding stage is carried out using the sine and cosine functions which aim to encode sequential information in time series data by adding element-wised from the input vector to the positional encoding vector. The resulting vectors are then fed into the four encoder layers. Each encoder layer consists of two sub-layers, namely the self-attention sub-layer and the fully-connected feed-forward sub-layer. Each sub-layer is followed by a normalization layer. The encoder generates ${d}$ dimensional vector to proceed to the decoder~\cite{Wu2020}.

\subsubsection{Decoder}
The decoder consists of an input layer, one or several identical decoder layers, and an output layer. The input to the decoder starts from the last data point on the input of the encoder. At the input layer, the decoder input is mapped into a dimensional vector. In addition to the two sub-layers in each encoder layer, a third sub-layer is inserted in the decoder, namely the self-attention mechanism at the encoder output. Finally, there is an output layer which maps the output of the last decoder layer to the targeted time sequence. The decoder also uses look-ahead masking and a one-position offset between the decoder input and the target output in the decoder to ensure that the prediction of the time series data point only depends on the previous data point~\cite{Wu2020}.

\subsubsection{Positional Encoding}
Positional Encoding (PE) is entering some relative or absolute position information of each value in the sequence~\cite{Vaswani2017}. In this model, PE is added to the Pre-Layer and Post-Layer input embedding at the bottom of the encoder and decoder stack. PE has the same dimensions as the embedding, so they can be added together~\cite{Wu2020}.

\subsubsection{Multi Head Attention}
Attention is a function that maps a query and a set of key-value pairs to output. Query, key, value, and output are in vector form. The output value is calculated as the sum of the weights of the values. The weight for each value is calculated by a query function with the corresponding key~\cite{Vaswani2017,Li2019,Lim2021}.

Multi Head Attention is a module on the attention mechanism that calculates the Attention function several times in parallel. The output results from each head are combined by adding up the matrix values (concat) of each and linearly converted into the expected dimensions~\cite{Vaswani2017,Li2019,Lim2021}.

Based on the Deep Transformer model by Wu et al~\cite{Wu2020}. There are three types of Multi Head Attention, those are Multi-Head Self-Attention (MHSA), Multi-Head Causal-Attention (MHCA), and Encoder-Decoder Attention. MHCA is different from MHSA, where MHCA will do masking for the next prediction result, unlike MHSA where the value for each data is known. Then, in the Encoder-Decoder Attention key and value are obtained from the encoder results while the query is obtained from the MHCA results.

\subsubsection{Residual Connection and Layer Normalization}
In Figure~\ref{architecture}, it can be seen that between the sub-layers of the encoder and decoder layers there is an Add and Normalize layer~\cite{Wu2020}. The add layer is the residual connection layer~\cite{He2015}, which is a skip-connection that learns the residual function by referring to the input layer, rather than studying the unreferenced function.

The normalization layer is designed to overcome the disadvantages of batch normalization~\cite{Ioffe2015} such as computationally expensive and batch size sensitive performance. Unlike batch normalization, layer normalization is a normalization method that directly estimates the normalization statistics from the inputs added to the neurons in the hidden layer so that normalization does not create new dependencies between training cases. The normalization layer can be calculated using Equation~\ref{eq:ln}, where ${E[x]}$ and ${Var[x]}$ are obtained from Equation~\ref{eq:ln_mu} and Equation~\ref{eq:ln_sigma}. Variable H indicates the number of hidden units in the layer.

\begin{equation}
\label{eq:ln}
\hat{x} = \frac{x - E[x]}{\sqrt{Var[x] + \epsilon}}
\end{equation}

\begin{equation}
\label{eq:ln_mu}
\mu^l = \frac{1}{H} \sum_{i=1}^{H} a_i^l
\end{equation}

\begin{equation}
\label{eq:ln_sigma}
\sigma^l = \sqrt{\frac{1}{H} \sum_{i=1}^{H} (a_i^l - \mu^l)^2}
\end{equation}

\subsubsection{Layer Normalization in Deep Transformer}
\begin{figure}
\centering
\includegraphics[width=6cm]{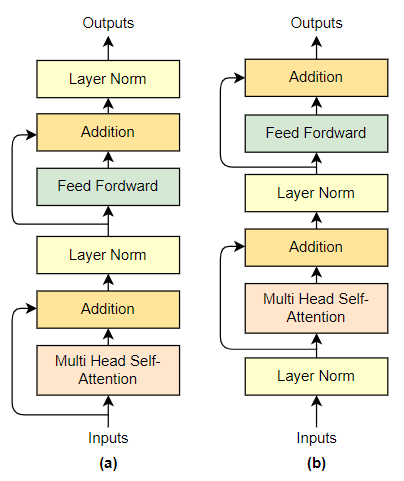}
\caption{(a) Post-LN (original Transformer); (b) Pre-LN.} \label{layernorm}
\end{figure}

Based on research by Baevski \& Auli, Child et al., Wang et al., and Xiong et al.~\cite{Baevski2018,Child2019,Wang2019,Xiong2020}, the original architecture of the Transformer model uses Post-Layer Normalization (Post-LN), where the normalization layer is placed between the residual connections, while in Pre-Layer Normalization (Pre-LN), the normalization layer is placed inside the residual connection (see Fig.~\ref{layernorm}).

Xiong et al.~\cite{Xiong2020} compared the performance of Pre-LN and Post-LN on language translation testing and found that the BLEU score and validation loss of the Pre-LN Transformer were better than post-LN. Furthermore, it was also found that Pre-LN converges faster than Post-LN and also found that changes in the position of the normalization layer affect changes in the optimizer.

\subsubsection{Feed Fordward Networks}
Each layer on the encoder and decoder consists of fully connected feed-forward networks called Position-wise Feed-Forward Networks (FFN). Position-wise FFN is a network with a feed-forward layer consisting of two dense layers used in the last dimension, which means the same dense layer is used for each position item in an order called point-wise. Between the two dense layers, there is a ReLU activation function~\cite{Vaswani2017}.

\subsection{Recurrent Neural Networks}
Recurrent Neural Networks (RNN) is a neural network architecture that aims to recognize patterns from sequence data. RNN can be used on time series data, genomes, handwriting, images and even to a higher level can be used in speech recognition. Mathematically, the hidden state and output of the RNN can be written in the form of an equation, those are Equation~\ref{eq:rnn_ht} and Equation~\ref{eq:rnn_ot}.

\begin{equation}
\label{eq:rnn_ht}
H_t = \phi_t(X_t W_{xh} + H_{t - 1} W_{hh} + b_h)
\end{equation}

\begin{equation}
\label{eq:rnn_ot}
O_t = \phi_o(H_t W_{ho} + b_o)
\end{equation}

The hidden state and the input at time t, respectively, are assumed to be ${H_t}$ and ${X_t}$. Next, the weights will be used, those are ${W_{hh}}$ for the hidden state and ${W_{xh}}$ for input, while the bias will be ${b_h}$. The output at time t is assumed to be ${O_t}$ with a weight of ${W_{ho}}$ and a bias of ${b_o}$. Finally, the results of the calculations for both will be passed to the activation function ${\phi}$ such as logistic sigmoid or tanh.

\subsection{Long Short-Term Memory}
LSTM is a neural network that can handle vanishing gradient problems in RNN by using gated cells that can store information. In LSTM, there are three gates, namely input gate ${I_t}$ to read data into the cell, output gate ${O_t}$ to read cell entries, and forget gate ${F_t}$ to reset the contents of the cell. The three gates can be written into equations respectively in Equation~\ref{eq:lstm_it}, Equation~\ref{eq:lstm_ot}, and Equation~\ref{eq:lstm_ft}. The variables ${W_{xi}}$, ${W_{xo}}$, and ${W_{xf}}$ are the data weights of each gate. The variables ${W_{ho}}$, ${W_{ho}}$, and ${W_{hf}}$ are the hidden state weights of each gate. While the variables ${b_i}$, ${b_o}$, and ${b_f}$ are the bias. The calculation results from the three gates will be passed to the sigmoid activation function.

\begin{equation}
\label{eq:lstm_it}
I_t = \sigma(X_t W_{xi} + H_{t - 1} W_{hi} + b_i)
\end{equation}

\begin{equation}
\label{eq:lstm_ot}
O_t = \sigma(X_t W_{xo} + H_{t - 1} W_{ho} + b_o)
\end{equation}

\begin{equation}
\label{eq:lstm_ft}
F_t = \sigma(X_t W_{xf} + H_{t - 1} W_{hf} + b_f )
\end{equation}

Next, there is a cell ${\tilde{C_t}}$ as a memory candidate which has the same calculation as before, but uses the activation function tanh. the calculation of ${\tilde{C_t}}$ will use the weights ${W_{xc}}$ and ${W_{hc}}$ as well as the bias ${b_c}$ which can be seen in Equation~\ref{eq:lstm_ct_tilde}. Finally, the value of the new memory will be calculated, namely ${C_t}$ and ${H_t}$ which can be seen in Equation~\ref{eq:lstm_ct} and Equation~\ref{eq:lstm_ht}, where ${\odot}$ denotes element-wise multiplication

\begin{equation}
\label{eq:lstm_ct_tilde}
\tilde{C_t} = tanh(X_t W_{xc} + H_{t - 1} W_{hc} + b_c)
\end{equation}

\begin{equation}
\label{eq:lstm_ct}
C_t = F_t \odot C_{t - 1} + I_t \odot \tilde{C_t}
\end{equation}

\begin{equation}
\label{eq:lstm_ht}
H_t = O_t \odot tanh(C_t)
\end{equation}

\subsection{Mean Absolute Percentage Error}
According to Adhikari and Agrawal~\cite{Adhikari2013}, Mean Absolute Percentage Error (MAPE) is a performance measure to measure the performance of a time series prediction or forecasting model. It can be seen in Equation~\ref{eq:mape}, MAPE represents the percentage of the average absolute error. The variable e is the error value of the prediction, y is the actual value, and n is the measure of the test data. The error value is obtained from the difference between the predicted value and the actual value.
\begin{equation}
\label{eq:mape}
MAPE=\frac{1}{n}\sum_{t=1}^{n}\left|\frac{e_t}{y_t}\right|*100
\end{equation}

\subsection{Data}
This study uses secondary data as the object of research. The data was taken from the public API on the official page of the COVID-19 Handling Task Force in Indonesia, that is covid.go.id. The data taken is in the form of a JSON (JavaScript Object Notation) file containing overall information on COVID-19 cases, daily updates, and accumulation from the beginning of the case to the present.

The data will be normalized using min-max normalization with a value range of -1 to 1. The data is then divided into three, those are training data, evaluation data, and test data. Test data is the data used to test the model that has been trained. The test data were taken from the last 60 days of the total data. Training data is the data used to train the model and is determined as much as 70\% of the data from the beginning except for test data. Evaluation data is used to evaluate the model in each epoch and is determined as much as 30\% of the data from the back except for test data.

\section{Experimental Results}
The test was carried out on 750 initial COVID-19 data in Indonesia with the last 60 days used as test data. First, testing is carried out to obtain optimal hyperparameters. The hyperparameters are: embedding size; the number of encoder and decoder blocks; FFN, pre-layer, and post-layer dimensions; and the amount of time lag. Furthermore, further testing is carried out, those are testing the number of predictions for the next day, optimizer used in training, the number of features, and finally comparing the Deep Transformer model with the LSTM and RNN models.

\subsection{Hyperparameter Testing}
Prior to testing to obtain optimal hyperparameters, the initial hyperparameters are set as benchmarks or defaults, that is: input step or time lag of 7; predict step or the number of predictions ahead of 1; $d$ size of 64; epoch maximum of 300; the number of encoders and decoders is 2 each; the number of heads from each encoder and decoder is 1; there are 50 hidden layers in feed forward, pre-layer, and post-layer each; drop out on pre-layer, post-layer, and model is 0.2; the position of the normalized layer is Pre-LN; At the beginning of the training the learning rate is determined to be 1 and will decrease as the training is carried out. The optimizer used is the Adam optimizer with the LambdaLR scheduler with a function created by Wu et al.~\cite{Wu2020}.

\subsubsection{Testing on Embedding Size}
The first test was carried out to obtain the best $d$ size among several sizes, those are 32, 64, 128, and 258. Table~\ref{tab:dmodel} is the result of testing the model on the $d$ size. In the MAPE column there are 3 sub-columns, those are Pre-LN, post-LN, and average. By default the Deep Transformer model uses the post-LN normalization layer type. The average column is the average of the test results for the two types of normalization layers.

\begin{table}
\begin{center}
\caption{The results of model testing on the size of $d_{model}$.}\label{tab:dmodel}
\setlength{\tabcolsep}{1em} 
{\renewcommand{\arraystretch}{1.3}
\begin{tabular}{|c|c|c|c|}
\hline
$d_{model}$ & \multicolumn{3}{c|}{MAPE} \\\cline{2-4} & Pre-LN & Post-LN & Mean\\
\hline
32	&20.07	&21.10	&20.585\\
64	&\bfseries{19.07}	&\bfseries{20.12}	&\bfseries{19.595}\\
128	&20.08	&20.46	&20.27\\
256	&19.92	&19.73	&19.825\\
\hline
\end{tabular}
}
\end{center}
\end{table}

Based on these test results, it was found that the best MAPE values for both types of normalized layers were equally obtained from the $d$ size of 64. Therefore, the $d$ size of 64 was used for advanced testing.

\subsubsection{Testing on the Number of Encoder and Decoder Blocks}
Furthermore, testing is carried out to obtain the best combination of the number of encoder and decoder blocks. The combination of the number of encoder and decoder blocks are 1-1, 2-2, 4-4, 2-4, and 4-2. Table~\ref{tab:encdec} shows the result of the test.

\begin{table}
\begin{center}
\caption{The results of testing the combination of the number of encoder and decoder blocks.}\label{tab:encdec}
\setlength{\tabcolsep}{1em} 
{\renewcommand{\arraystretch}{1.3}
\begin{tabular}{|c|c|c|c|c|}
\hline
Encoder & Decoder & \multicolumn{3}{c|}{MAPE}\\
\cline{3-5} & & Pre-LN & Post-LN & Mean\\
\hline
1	&1	&20.52	&20.93	&20.725\\
2	&2	&\bfseries{19.07}	&\bfseries{20.12}	&\bfseries{19.595}\\
4	&4	&21.23	&21.36	&21.295\\
2	&4	&20.28	&20.59	&20.435\\
4	&2	&21.08	&20.82	&20.95\\
\hline
\end{tabular}
}
\end{center}
\end{table}

Based on the results of these tests, it was found that the combination of the number of encoder and decoder blocks which both amounted to 2 got the best MAPE value for the second type of normalization layer. Therefore, the number of encoder and decoder blocks used for advanced testing is 2.

\subsubsection{Testing the Dimensions of the Feed Forward Network, Pre-Layer, and Post-Layer}
The next test combines the dimensions of the Feed Fordward Network layer, Pre-Layer, and Post-Layer. The size dimensions are either 30, 50, or 100. Table~\ref{tab:dims} shows the result of the test.

\begin{table}
\begin{center}
\caption{The test results for the combination of FFN, Pre-Layer, and Post-Layer dimensions.}\label{tab:dims}
\setlength{\tabcolsep}{1em} 
{\renewcommand{\arraystretch}{1.3}
\begin{tabular}{|c|c|c|c|c|c|}
\hline
FFN & Pre-Layer & Post-Layer & \multicolumn{3}{c|}{MAPE}\\
\cline{4-6} & & & Pre-LN & Post-LN & Mean\\
\hline
30	&30	    &30	    &20.04	&20.90	&20.47\\
50	&50	    &50	    &19.07	&20.12	&19.60\\
100	&100	&100	&19.87	&21.44	&20.66\\
100	&50	    &50	    &18.81	&\bfseries{20.06}	&\bfseries{19.44}\\
50	&100	&50	    &\bfseries{18.70}	&20.34	&19.52\\
50	&50	    &100	&21.71	&21.71	&21.71\\
\hline
\end{tabular}
}
\end{center}
\end{table}

Based on the test results, it was found that the Pre-Layer with dimensions of 100 and the rest with dimensions of 50 got the best MAPE value for the Pre-LN model. Meanwhile, the best post-LN model yields the best performance using dimensions of FFN of 100 and the rest has dimensions of 50. Then the average is calculated for models with both normalization layers to determine which combination will be used for forthcoming testing. Finally, the best average is obtained for the FFN with the dimension of 100 and the remaining of 50.

\subsubsection{Testing the Amount of Time Lag}
Next, the number of time lags was tested, which were 4, 7, 14, and 30 days. Table~\ref{tab:timelag} is the result of testing the amount of time lag. Based on these results, it was found that the 7-day time lag got the best MAPE value for both types of normalized layers. Therefore, a time lag of 7 days was used for further testing.

\begin{table}
\begin{center}
\caption{The test results for the combination of FFN, Pre-Layer, and Post-Layer dimensions.}\label{tab:timelag}
\setlength{\tabcolsep}{1em} 
{\renewcommand{\arraystretch}{1.3}
\begin{tabular}{|c|c|c|c|}
\hline
Time Lag & \multicolumn{3}{c|}{MAPE}\\
\cline{2-4} & Pre-LN & Post-LN & Mean\\
\hline
4	&19.59	&20.17	&19.88\\
7	&\bfseries{19.07}	&\bfseries{20.12}	&\bfseries{19.595}\\
14	&24.42	&25.28	&24.85\\
30	&24.16	&25.8	&24.98\\
\hline
\end{tabular}
}
\end{center}
\end{table}

\subsection{Testing and Analysis of the Number of Predicted Days in the Future}
\begin{figure}[H]
\includegraphics[width=\textwidth]{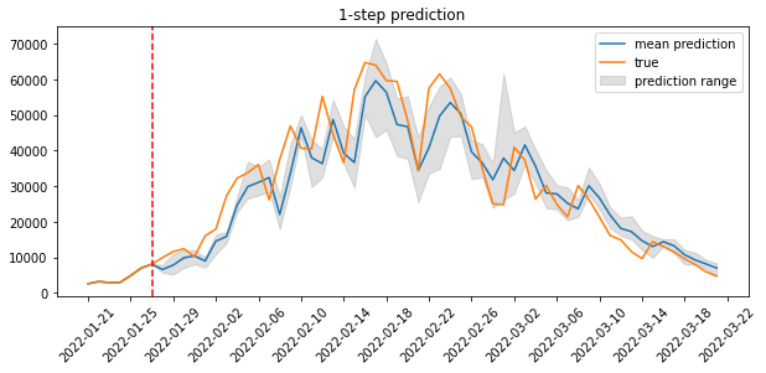}
\caption{Prediction results 1 day ahead.} \label{pred1}
\end{figure}

After obtaining the best hyperparameters from the Deep Transformer model, further testing is carried out, testing the number of predictions for the next day. The test is performed using the $d$ size of 64, the number of encoder and decoder blocks of 2 each, FFN dimension of 100, Pre-Layer and Post-Layer dimensions of 50, and the amount of time lag of 7 days. Then, judging from the previous four tests, the Pre-LN normalization layer got a better error value than the post-LN. Therefore, Pre-LN will be used for further testing.
First, testing is carried out for predictions one day ahead of the last 60 days. The test was carried out ten times in a row and the average with its standard deviation of the error was calculated. The average MAPE value obtained is 22.298 and the standard deviation is 1.363. Figure~\ref{pred1} shows a graph comparing the average prediction results to the original value along with the range of prediction results from 10 trials.

\begin{figure}[H]
\includegraphics[width=\textwidth]{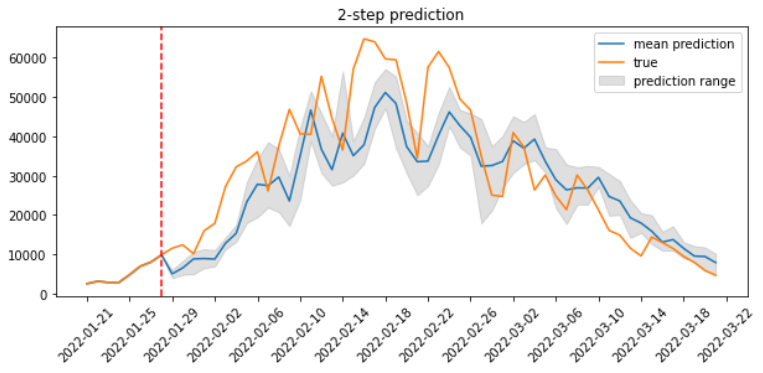}
\caption{Prediction results 2 day ahead.} \label{pred2}
\end{figure}

Second, testing is carried out for the next two days with the same data and method as the previous test. These tests obtained an average of 31.15 and a standard deviation of 2.901. The comparison of the average prediction results to the original value can be seen in Figure~\ref{pred2}.

\begin{figure}[H]
\includegraphics[width=\textwidth]{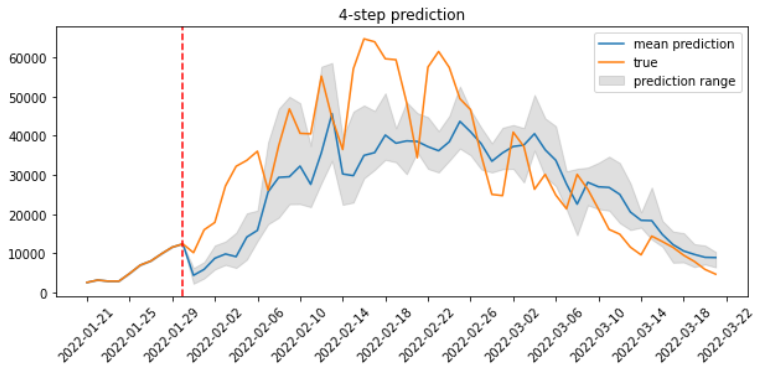}
\caption{Prediction results 4 day ahead.} \label{pred3}
\end{figure}

Next, testing is carried out for the next four days with the same data and method as the previous test. These tests obtained an average of 37.428 and a standard deviation of 1.635. The comparison of the average prediction results to the original value can be seen in Figure~\ref{pred3}.

\begin{figure}[H]
\includegraphics[width=\textwidth]{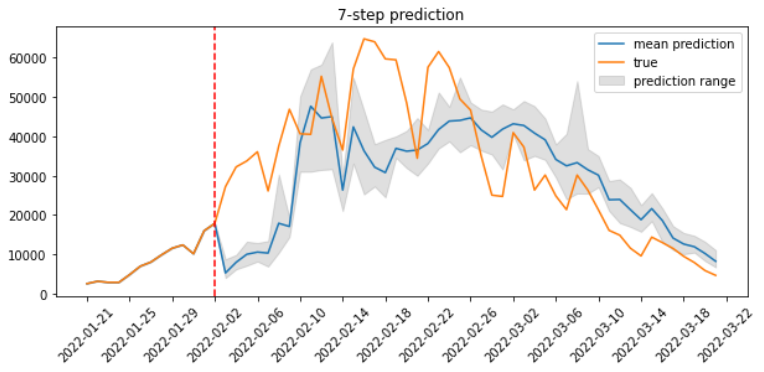}
\caption{Prediction results 7 day ahead.} \label{pred4}
\end{figure}

Furthermore, testing is carried out for the next 7 days with the same data and method as the previous test. From these tests, obtained an average of 42.35 and a standard deviation of 2.893. The comparison of the average prediction results to the original value can be seen in Figure~\ref{pred4}.

Based on the results of the four tests, it can be seen that the prediction error value for the next 1 day is the smallest and predictions for the next 2, 4, and 7 days will increase along with the increase in the number of predictions for the next day. Therefore, it can be said that the more days that will be predicted in the future, the resulting error value will increase. The comparison of the average error value and standard deviation can be seen in Table~\ref{tab:ndayspred}.

\begin{table}[H]
\begin{center}
\caption{Test results on the number of predictions for the next n-days.}\label{tab:ndayspred}
\setlength{\tabcolsep}{1em} 
{\renewcommand{\arraystretch}{1.3}
\begin{tabular}{|c|c|c|}
\hline
N-day prediction & \multicolumn{2}{c|}{MAPE}\\
\cline{2-3} & Mean & Standard Deviation\\
\hline
1	&\bfseries{22.298}	&\bfseries{1.363}\\
2	&31.150	&2.901\\
4	&37.428	&1.635\\
7	&42.350	&2.893\\
\hline
\end{tabular}
}
\end{center}
\end{table}

\subsection{Testing and Analysis of the Optimizer Used}
Furthermore, research was conducted on the optimizer used in training. The optimizers tested are Adam, AdamW, Adamax, Adagrad, Adadelta, SGD, and RMSprop. The hyperparameters used and the method of testing is the same as testing the predictions for the next day. For each optimizer, a LambdaLR scheduler is used with a function created by Wu et al.~\cite{Wu2020}.

\begin{figure}[H]\centering
\includegraphics[width=\textwidth]{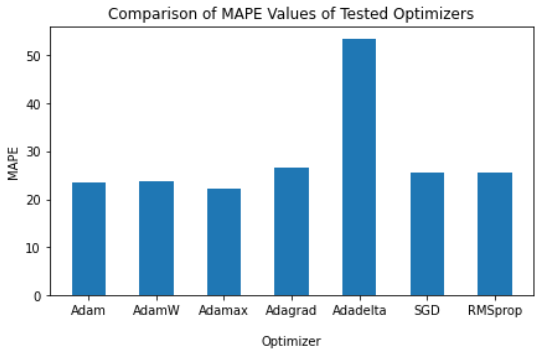}
\caption{Comparison of MAPE values of tested optimizers.} \label{optimizer_comparison}
\end{figure}

\begin{figure}[H]
\includegraphics[width=\textwidth]{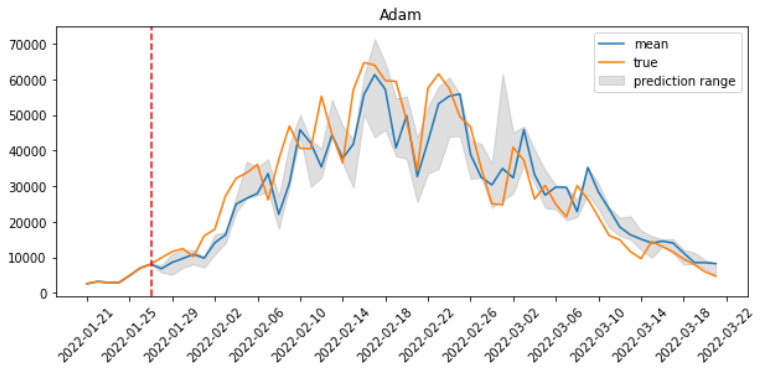}
\caption{Prediction results for optimizer Adam.} \label{adam}
\end{figure}

\begin{figure}
\includegraphics[width=\textwidth]{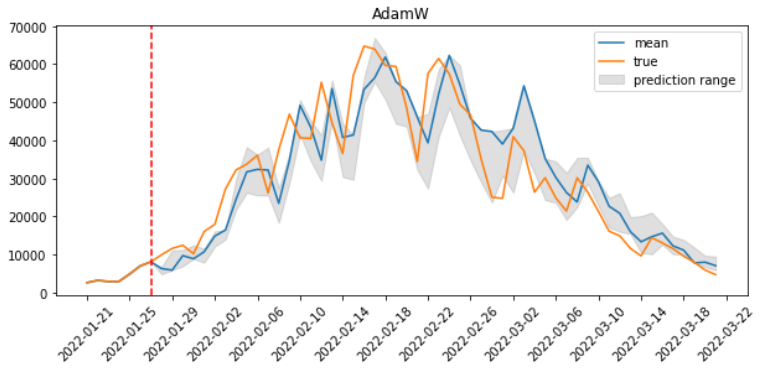}
\caption{Prediction results for optimizer AdamW.} \label{adamw}
\end{figure}

\begin{figure}
\includegraphics[width=\textwidth]{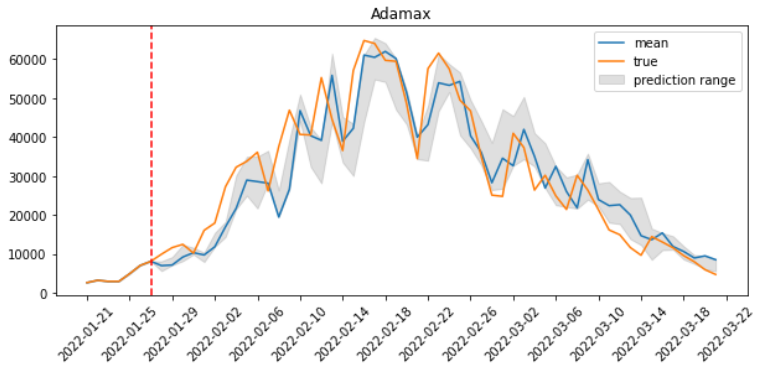}
\caption{Prediction results for optimizer Adamax.} \label{adamax}
\end{figure}

\begin{figure}
\includegraphics[width=\textwidth]{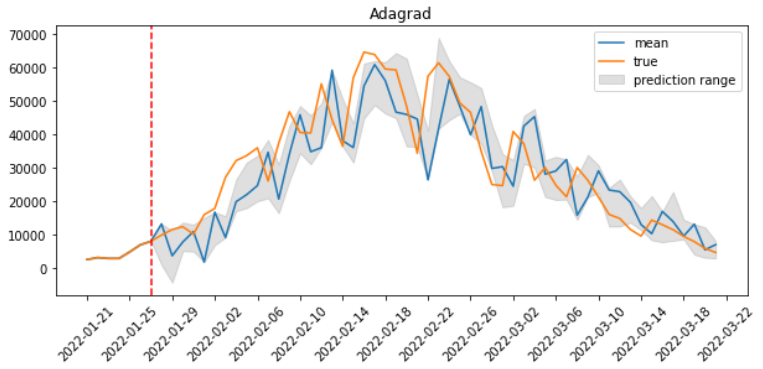}
\caption{Prediction results for optimizer Adagrad.} \label{adagrad}
\end{figure}

\begin{figure}
\includegraphics[width=\textwidth]{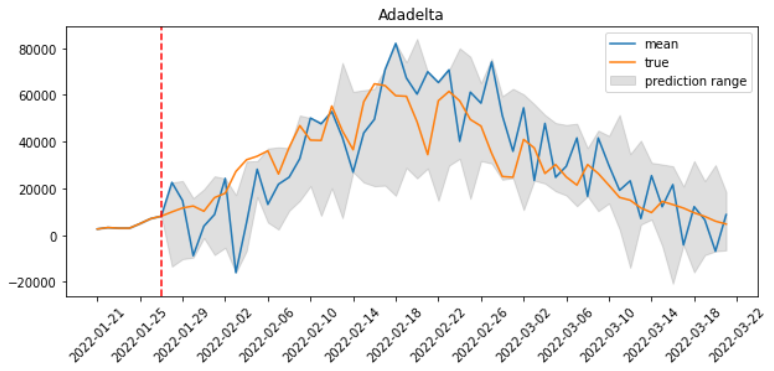}
\caption{Prediction results for optimizer Adadelta.} \label{adadelta}
\end{figure}

\begin{figure}
\includegraphics[width=\textwidth]{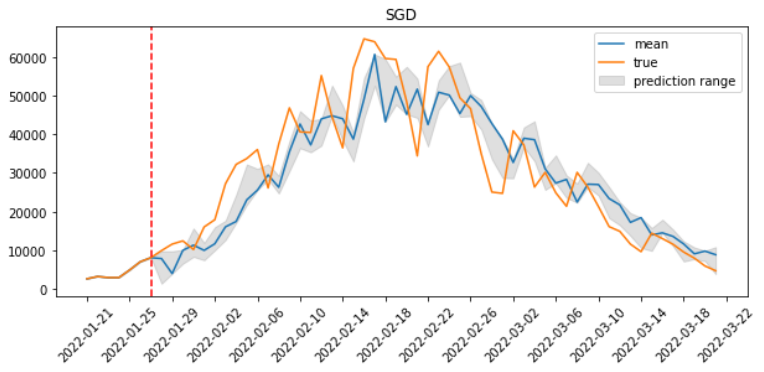}
\caption{Prediction results for optimizer SGD.} \label{sgd}
\end{figure}

\begin{figure}
\includegraphics[width=\textwidth]{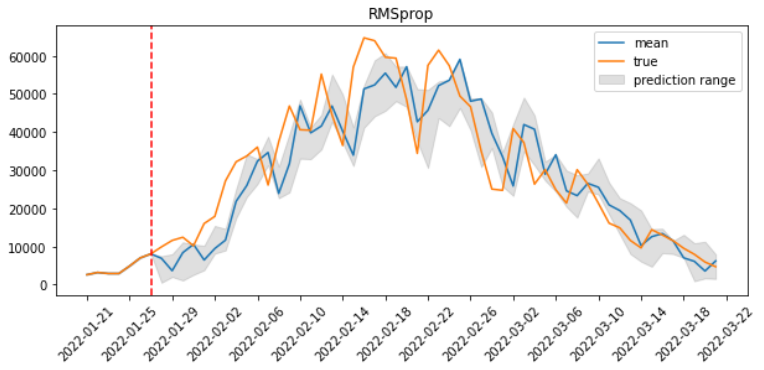}
\caption{Prediction results for optimizer RMSprop.} \label{rmsprop}
\end{figure}

The results of comparing the MAPE values of the tested optimizer can be seen in Figure~\ref{optimizer_comparison}. In contrast, the graph of the comparison of the average prediction results with the ground-truth along with the range of the prediction results can be seen in Figure~\ref{adam}, Figure~\ref{adamw}, Figure~\ref{adamax}, Figure~\ref{adagrad}, Figure~\ref{adadelta}, Figure~\ref{sgd}, and Figure~\ref{rmsprop}.

Based on the results of the seven tests, it can be seen that the prediction error value using the Adamax optimizer is the smallest followed by Adam, AdamW, SGD, Adagrad, RMSprop, and the worst is the Adadelta optimizer. Therefore, it can be said that the Adamax optimizer can train the model better with a MAPE value of 22.343 compared to other optimizers. The comparison of the average error value and standard deviation can be seen in Table~\ref{tab:optimizer}.

\begin{table}[H]
\begin{center}
\caption{Test results on the optimizer used.}\label{tab:optimizer}
\setlength{\tabcolsep}{1em} 
{\renewcommand{\arraystretch}{1.3}
\begin{tabular}{|c|c|c|}
\hline
Optimizer & \multicolumn{2}{c|}{MAPE}\\
\cline{2-3} & Mean & Standard Deviation\\
\hline
Adam	    &23.415	&1.846\\
AdamW	    &23.870	&1.825\\
Adamax	    &\bfseries{22.343}	&1.764\\
Adagrad	    &26.560	&3.163\\
Adadelta	&53.410	&5.812\\
SGD	        &25.510	&\bfseries{1.708}\\
RMSprop	    &26.700	&2.501\\
\hline
\end{tabular}
}
\end{center}
\end{table}

\subsection{Testing and Analysis of the Number of Features}
Furthermore, the number of features is tested, those are 1 feature (univariate) and multivariate with 3 features. One feature is using the number of positive cases, while three features are the number of positive cases, the number of deaths, and the number of recoveries. Both tests have one output, that is the number of positive cases. The hyperparameters used are the same as the previous advanced tests. Testing is done by repeating the training and testing process several times for the last 60 days of data until the best results are obtained.

\begin{figure}[H]
\includegraphics[width=\textwidth]{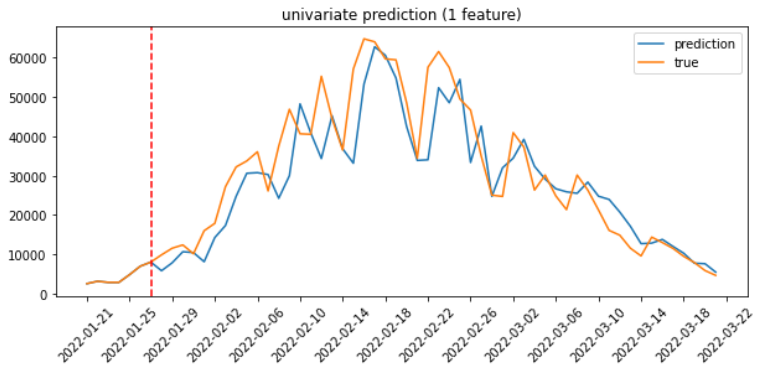}
\caption{Prediction results with 1 feature.} \label{univariate}
\end{figure}

\begin{figure}[H]
\includegraphics[width=\textwidth]{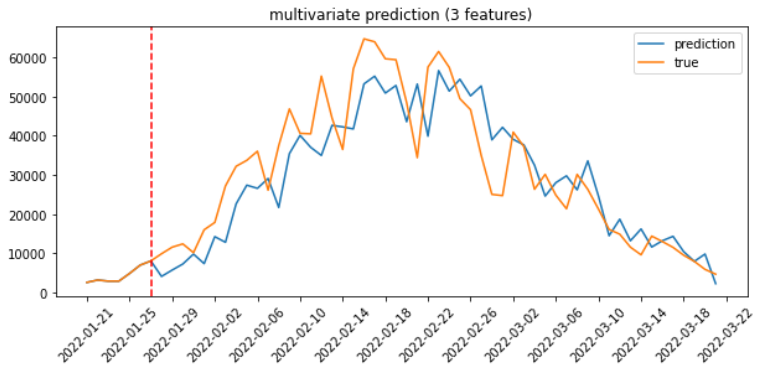}
\caption{Prediction results with 3 features.} \label{multivariate}
\end{figure}

After several iterations of testing, the best result is that the model with one feature gets a MAPE value of 18.83, while the three features get a MAPE value of 25.24. The comparison graph between the prediction results and the original value of the prediction of one feature and 3 features, respectively, can be seen in Figure~\ref{univariate} and Figure~\ref{multivariate}.

\subsection{Testing and Analysis of Other Models}
Finally, a test was conducted to compare the Deep Transformer model with other models, those are Long Short-Term Memory (LSTM) and Recurrent Neural Network (RNN). The Deep Transformer model uses hyperparameters as before with a Pre-LN normalization layer, while the LSTM and RNN models use 16 hidden layers, one layer, 2000 epochs, and use the Adam optimizer with a learning rate of 0.01. Testing is carried out by conducting the training and testing process several times for the data for the last 60 days until the best results are obtained.

\begin{figure}[H]
\centering
\includegraphics[width=7cm]{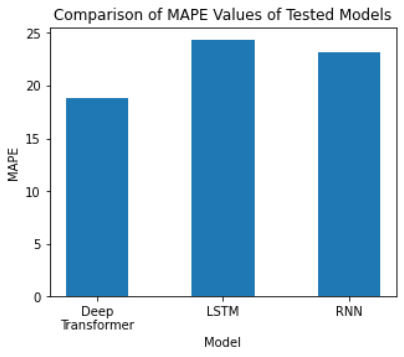}
\caption{Comparison of MAPE values of Deep Transformer, LSTM, and RNN.} \label{model_comparison}
\end{figure}

\begin{figure}[H]
\includegraphics[width=\textwidth]{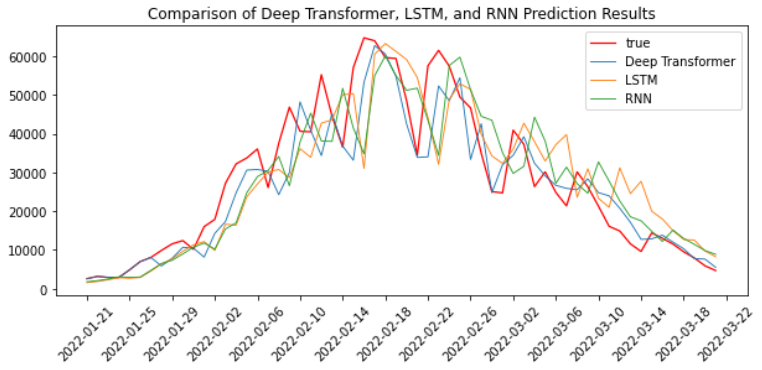}
\caption{Comparison of the prediction results of Deep Transformer, LSTM, and RNN.} \label{other_models}
\end{figure}

After several training and testing processes were carried out, the best MAPE results were obtained for each model. The Deep Transformer model got the best MAPE result of 18.83, the LSTM model got the MAPE result of 24.33, while the RNN model got the MAPE result of 23.15. A more detailed comparison of the MAPE values of each model can be seen in Figure~\ref{model_comparison}, while a comparison of the prediction results for each model can be seen in Figure~\ref{other_models}.

Based on the test results of the three models, it can be said that the Deep Transformer model outperforms the other two models by a considerable difference. The performance of the LSTM and RNN models is almost the same, but the RNN model is superior to LSTM with the same number of hidden layers.

\section{Conclusion}
The design of the Deep Transformer model in predicting the growth of COVID-19 cases in Indonesia starts by taking datasets from the public API on the official page of the COVID-19 Handling Task Force in Indonesia. Then, from the dataset, the required column is taken. Those are the date and the positive number of 750 days since the beginning of the case. Furthermore, the Deep Transformer model was developed in such a way that it could be used to predict the number of positive cases from the COVID-19 dataset in Indonesia. After the model is completed, training is carried out to obtain a model that can optimally predict positive cases by testing several existing hyperparameters. After obtaining the best hyperparameters, the testing process is carried out to predict the number of positive cases from the last 60 days of data. Finally, the test results will calculate the error value using Mean Absolute Percentage Error (MAPE).

Based on the results of several tests, the best hyperparameters were obtained. These are: embedding size of 64, the number of encoder and decoder blocks of two each, dimensions of the FFN of 100, Pre-Layer and Post-Layer dimensions are 50, the amount of time lag of 7 days, drop out on Pre-Layer of 0.2, the normalized layer of Pre-LN, and using the Adam optimizer of Adam with a learning rate initialization of 1. The loss will decrease as the training progresses with a total of 300 epochs. Using these hyperparameters, the Deep Transformer model gets a MAPE value of 18.83.

Suggestions for further experiments are comparing the results of model training with several optimizers other than Adam. Furthermore, developing a proper Deep Transformer model with more than one feature or multivariate input that can predict more than one day is promising.

%
%
%
\bibliographystyle{splncs04}
\bibliography{dafpus.bib}

\end{document}